\documentclass[10pt,twocolumn,letterpaper]{article}

\usepackage{cvpr}
\usepackage{times}
\usepackage{epsfig}
\usepackage{graphicx}
\usepackage{amsmath}
\usepackage{amssymb}
\usepackage{multirow}
\usepackage{float}
\usepackage{subfigure}
\usepackage{color}
\usepackage{array}
\usepackage{tabularx}

\usepackage[breaklinks=true,bookmarks=false]{hyperref}

\cvprfinalcopy 

\def\cvprPaperID{686} 

\begin{document}
	
	\title{Deep RNN Framework for Visual Sequential Applications}
	\author{Bo Pang\thanks{Equal contribution.} \quad\ Kaiwen Zha\footnotemark[1] \quad\ Hanwen Cao \quad Chen Shi \quad Cewu Lu\thanks{Corresponding author. Cewu Lu is a member of Department of Computer Science and Engineering, Shanghai Jiao Tong University, MoE Key Lab of Artificial Intelligence, AI Institute, Shanghai Jiao Tong University.}\\
	Shanghai Jiao Tong University\\
	{\tt \small \{pangbo, kevin\_zha, mbd\_chw, shichen, lucewu\}@sjtu.edu.cn}}

	\maketitle
	\pagestyle{empty}
	\thispagestyle{empty}
	
	\begin{abstract}
	Extracting temporal and representation features efficiently plays a pivotal role in understanding visual sequence information. To deal with this, we propose a new recurrent neural framework that can be stacked deep effectively. There are mainly two novel designs in our deep RNN framework: one is a new RNN module called Context Bridge Module (CBM) which splits the information flowing along the sequence (temporal direction) and along depth (spatial representation direction), making it easier to train when building deep by balancing these two directions; the other is the Overlap Coherence Training Scheme that reduces the training complexity for long visual sequential tasks on account of the limitation of computing resources.
	
	We provide empirical evidence to show that our deep RNN framework is easy to optimize and can gain accuracy from the increased depth on several visual sequence problems. On these tasks, we evaluate our deep RNN framework with 15 layers, 7$\times$ than conventional RNN networks, but it is still easy to train. Our deep framework achieves more than 11\% relative improvements over shallow RNN models on Kinetics, UCF-101, and HMDB-51 for video classification. For auxiliary annotation, after replacing the shallow RNN part of Polygon-RNN with our 15-layer deep CBM, the performance improves by 14.7\%. For video future prediction, our deep RNN improves the state-of-the-art shallow model's performance by 2.4\% on PSNR and SSIM. The code and trained models are published accompanied by this paper: \url{https://github.com/BoPang1996/Deep-RNN-Framework}.
    \end{abstract}

	\vspace{-0.1in}
	\section{Introduction}
	With the advent of deep neural networks (DNN) in recent years, a mass of vision tasks have made great progress~\cite{krizhevsky2012imagenet, zeiler2014visualizing,sermanet2013overfeat,redmon2018yolov3,he2017mask} due to its superior representation capability for high-dimensional data. 
	On top of spatial representation, temporal features are valuable and crucial as well when dealing with sequential inputs like videos, for which recurrent neural networks (RNN) are designed. Taking all above into consideration, we are intended to build a deep RNN architecture that combines the merits of both RNN and DNN to extract more powerful temporal and representation features from visual sequential inputs.
	
	A straightforward way to build RNN deeper is to simply stack multiple RNN layers. However, this method is encountered with two problems. For one thing, in this deep RNN structure, there exist two information flows --- \textit{representation flow} and \textit{temporal flow}, along structural (spatial) depth and temporal depth respectively, however, these two flows are often entangled with each other, making it hard for models to be co-adaptive to both of them. Many specific RNN structures like LSTM~\cite{hochreiter1997long} and GRU~\cite{cho2014learning} are designed mainly to capture temporal information among long sequences, yet there is no adaption that can effectively take advantage of both the two flows. Therefore, simply stacking these RNN modules will lead to higher training error and heavier training consumption. For another, the limitation of computing resources greatly influences the feasibility of this method. Unlike deep CNN~\cite{Pan_2018_CVPR,li2018crowdpose,li2018transferable}, deep RNN needs to unfold as many times as the sequence length, resulting in more significant expansion of memory and computational complexity with the depth increasing, especially for visual sequential inputs. 
	
	
	In this paper, we propose a new deep RNN architecture including two principle techniques, namely, \textit{Context Bridge Module} (CBM) and \emph{Overlap Coherence Training Scheme}. In CBM, we design two computing units taking charge of representation flow and temporal flow respectively, forcing these two flows relatively independent of each other with the aim of making them focus on representation and temporal information separately to ease the training process. After these two units, a merge unit is utilized to synthesize them. By adjusting the synthesizing method, we can balance the dominant degree of each direction to better adapt to the requirements of different tasks. Furthermore, to make representation flow less influenced by temporal flow in the beginning of training, we design the \emph{Temporal Dropout} (TD) to interdict the back-propagation of temporal information across layers with a certain probability.
	
	
	Besides, the proposed \emph{Overlap Coherence Training Scheme} aims at reducing the training cost of deep RNN. Since the enormous training consumptions are largely due to the long sequence, we introduce this training scheme that randomly samples the long sequence with length $l$ into several overlapping short clips with length $n$ and leverages the overlaps as the communication bridge between the adjacent clips to smooth the information propagation among clips. In this way, we simplify the original Markov process of order $l$ into several ones of order $n$ ($n < l$), which remarkably reduces the training complexity, and guarantees the temporal information coherence among clips at the same time. Based on overlaps, we design \textit{overlap coherence loss} that forces the detached clips to generate coherent results in order to strengthen the consistency of temporal information, which makes the model not a strict Markov process of order $n$, but the complexity is still reduced.
	
	%
	
	
	We conduct comprehensive experiments on several tasks to show the challenge of training deep RNN and evaluate our proposed deep RNN framework. Results reveal that: 1) Deep RNN can enjoy accuracy gains from the greatly increased depth, substantially better than the shallow networks. 2) Our CBM is more suitable for stacking deep compared with other RNN structures like LSTM. 3) The overlap coherence training scheme can effectively make many computer vision problems with high-dimensional sequential inputs trainable on commonly-used computing devices.
	
	We evaluate our framework on several visual sequence tasks: action recognition and anticipation on UCF-101~\cite{soomro2012ucf101}, HMDB-51~\cite{kuehne2011hmdb} and Kinetics~\cite{carreira2017quo}, auxiliary annotation (Polygon-RNN~\cite{castrejon2017annotating}) on Cityscapes~\cite{cordts2016cityscapes}, and video future prediction on KTH~\cite{schuldt2004recognizing}. 
	For action recognition and anticipation tasks, our deep RNN framework achieves more than 11\% relative improvements on all the datasets compared with the shallow RNN models. For Polygon-RNN task, IoU value improves by 14.7\% on Cityscapes. For video future prediction task, our deep RNN improves the performance by 2.4\% on PSNR~\cite{mathieu2015psrn} and SSIM~\cite{wang2004ssim} metrics.
	%
	%
	
	\vspace{-0.05in}
	\section{Related Work}
	\paragraph{Methods for Visual Sequence Tasks}
	Visual sequence problems require models to extract hierarchical temporal and representation features simultaneously. A slew of prior arts have shed light on this tough problem: 1) An inchoate approach is pooling the spatial representation features of every item in the sequence, such as~\cite{karpathy2014large,yue2015beyond} when dealing with video classification and \cite{wang2016actionness,weinzaepfel2015learning} for action detection and localization. This approach can extract relative high-quality spatial representation features but is very weak for temporal ones because it treats the sequence as a set and simply combines the spatial features of the set as global temporal features without considering order relations. 2) Then 3D convolutional networks~\cite{ji20133d,carreira2017quo} appear, which treat temporal dimension equal to spatial dimension with its cubic convolution kernel, while 3D convolutional networks need to consume large amount of computing resources. 3) RNN~\cite{wu2015modeling,donahue2015long} is designed to handle sequence problems, therefore it is a natural idea to utilize RNN to encode temporal information after obtaining spatial features, which is adopted in~\cite{wu2015modeling,donahue2015long,li2018videolstm,pang2018human} for video classification, \cite{donahue2015long,venugopalan2014translating} for video description, \cite{castrejon2017annotating,acuna2018efficient} for auxiliary annotation and \cite{villegas2017decomposing,xingjian2015convolutional,oh2015action} for video future prediction. Whereas, currently used RNN is shallow, which may limit its performance.
	\vspace{-0.16in}
	\paragraph{Exploration on Deep RNN}
	In this paper, we focus on exploring appropriate deep structure for RNN model. There are many previous works trying to address this problem. In~\cite{pascanu2013construct,hermans2013training}, the authors evaluate several ways to extend RNN deeper, and results show that stacked RNN has relatively better performance and more importantly, stacking method can synthesize temporal information in each layer to extract hierarchical temporal-spatial features instead of plain temporal, deep spatial features.
	
	The learning computational complexity of deep RNN significantly increases with the depth growing, thus in~\cite{sak2014long}, the authors propose a new RNN structure called LSTMP to reduce the complexity. In ~\cite{irsoy2014deep,irsoy2014opinion,huang2015joint,hermans2013training}, researchers prove that deep RNNs outperform associated shallow counterparts that employ the same number of parameters. \cite{irsoy2014deep} shows that each layer captures a different aspect of compositionality which reveals deep RNN's ability to extract hierarchical features, and a deep bidirectional RNN structure is proposed in~\cite{irsoy2014opinion}. All these previous works prove the importance of RNN depth in NLP and speech area, while for high-dimensional inputs like videos in computer vision, it is more challenging to tackle as we mentioned above. For them, what we suppose to build is a deep RNN framework which is easy to optimize even when inputs are large-scale and can achieve promising improvements on performance at the same time.
	
	\vspace{-0.05in}
	\section{Deep RNN Framework}
	Deep model has exhibited superior performance in producing powerful features, and we hope sequence modeling can enjoy the deep representation as well. To this end, we introduce our deep RNN framework in this section, which contains two parts: context bridge module (CBM) designed to effectively capture temporal and representation information simultaneously, and the overlap coherence training scheme to further simplify the training process.
	
	\vspace{-0.05in}
	\subsection{Context Bridge Module}
	
	To model visual sequential inputs, we need to make sure it can be trained efficiently when building deep. For this, we design a non-shallow recurrent architecture to respectively capture temporal information from sequential inputs (e.g. a sequence of frames in a video) and representation information from each individual one (e.g. one frame of the sequence). These two information flows are oriented towards temporal depth and structural depth separately, and we name them as temporal flow and representation flow.
	
	\vspace{-0.15in}
	\paragraph{Challenge} The straight-forward design for deep RNN can be a vertically stacked RNN architecture. However, in high-dimensional visual tasks, parameters in RNN cell are hard to be co-adaptive to two flows simultaneously, resulting in ineffective and inefficient training. Extensive experiments show this design is very hard to train. This is why we hardly see stacked deep RNN in related literatures. In most cases, people adopt shallow RNN which takes extracted CNN features as inputs, though it is not an end-to-end pipeline.
	
	\begin{figure}
		\begin{center}
			\includegraphics[width=\linewidth, height=1.8in]{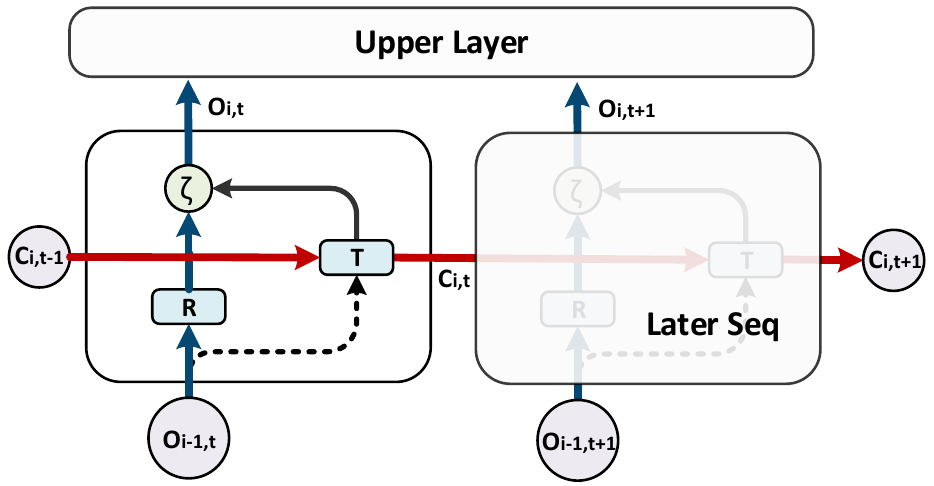}
		\end{center}
		\caption{Structure of CBM. The blue lines represent representation flows, while red ones represent temporal flows. $R,T$ and $\zeta$ denote representation unit, temporal unit and merge function respectively. The dashed line (TD) means feeding forward is allowed but back-propagation is forbidden with a certain probability.}
		\label{fig:DCFM}
		\vspace{-0.15in}
	\end{figure}
	
	\vspace{-0.15in}
	\paragraph{Our Architecture} Therefore, we go down to consider how to capture these two branches of information flows as independently as possible, through which the training process can be much easier since the two relatively independent branches can share the burden of learning and ease complex co-adaptations. Specifically, for representation flow, we use a computing unit (e.g. CNN structure) to extract features of the individual input sample without recurrent operations, while temporal flow adopts a RNN structure.
	
	As shown in Fig.~\ref{fig:DCFM}, in each cell, there is a ``representation" unit $R$ and a ``temporal" unit $T$ which act as a representation feature extractor on individual input sample and a temporal information encoder on the sequential inputs respectively. Here $R$ can be seen as a context bridge over the temporal information. Intuitively, the representation information flow would be encouraged to mainly propagate by this bridge, since it doesn't need to interwind with temporal information. Therefore, we call this module as \textit{Context Bridge Module} (CBM). By denoting $o_{i-1,t}$ as the input to the module in $i^{th}$ layer at time stamp $t$, we have
	\vspace{-0.05in}
	\begin{equation}
	o'_{i,t} = R(o_{i-1,t}; \psi_i),
	\end{equation} 
	where the representation unit $R$ is designed as a conventional CNN layer, namely $ReLU(Conv(\cdot))$, and $\psi_i$ is the parameters of $R$ in $i^{th}$ layer.
	
	On the other hand, temporal flow is captured by $T$ unit, which is written as
	\vspace{-0.08in}
	\begin{equation}
	c_{i, t} = T(c_{i, t-1}, o_{i-1,t}; \phi_i),
	\end{equation}
	where $c_{i, t}$ is the memory state in $i^{th}$ layer at time stamp $t$, and $\phi_i$ is the parameters of $T$ in $i^{th}$ layer. As a recurrent architecture, $T$ can be a $Sigmoid(Conv(\cdot))$ (as simple as the conventional RNN) or LSTM. In practice, we suggest $Sigmoid(Conv(\cdot))$ since it only consumes half of computing resources compared with LSTM cell, which greatly contributes to building model deeper.
	
	Finally, to fuse the information flows from the two units, we introduce a merge unit,
	\vspace{-0.08in}
	\begin{equation}
	o_{i, t} =  \zeta (o'_{i,t}, c_{i,t}),
	\end{equation}
	where $\zeta$ is the merge function, and we adopt element-wise production for $\zeta$ in our experiments.

	\begin{figure}
		\begin{center}
			\subfigure{
				\label{fig:detach:a} 
				\includegraphics[width=\linewidth]{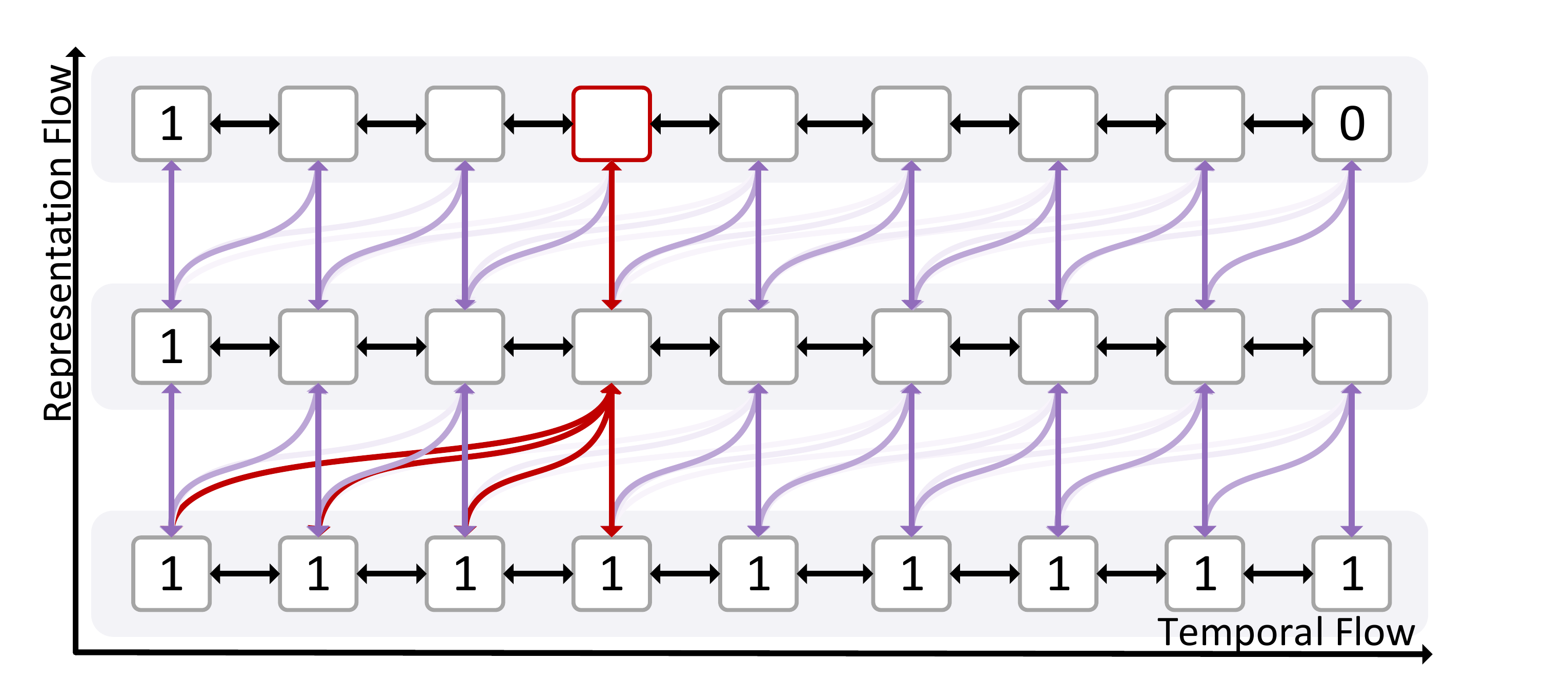}}
			\subfigure{
				\label{fig:detach:b} 
				\includegraphics[width=\linewidth]{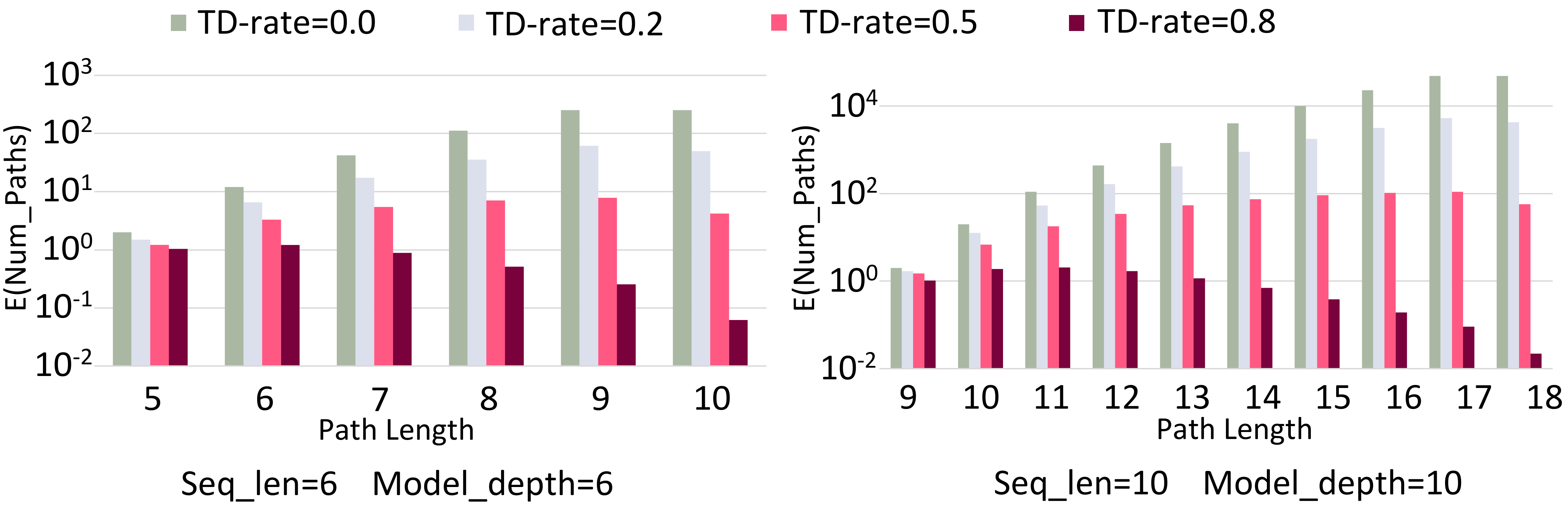}}
		\end{center}
		\caption{Temporal flows adopting TD. \textbf{Top:}  When setting the TD rate to 1.0, all the colorized lines (red \& purple) of temporal flow cannot propagate back, while if only drop the red node out, the gradients from red node's temporal unit cannot flow backward through the red lines. \textbf{Bottom:} Expectation numbers of back-propagation paths with different lengths (paths from ``0" to different ``1" in Top) when adopting different TD rates. Note that the back-propagation remains unchanged when setting TD rate to 0.0.}
		\vspace{-0.15in}
		\label{fig:detach}
	\end{figure}
	
	\vspace{-0.15in}
	\paragraph{Temporal Dropout}  To make training easier, we hope the learning in representation flow direction less interwinds with temporal flow in the beginning. After a desirable neural representation is shaped, the learning in temporal flow direction can be more efficient. To this end, we introduce a \emph{Temporal Dropout} (TD) scheme: forbidding back-propagation from $T$ unit through the dashed line in Fig.~\ref{fig:DCFM} with a certain probability. Just like dropout proposed in~\cite{krizhevsky2012imagenet}, it can reduce complex co-adaptations of two flows and enhance model's generalization ability. Specifically, we begin with a high temporal dropout rate (forbidding with a high probability) to isolate temporal information of each layer. In this way, the representation unit can capture effective representation easily, since it largely shortens the back-propagation chain in temporal flow as shown in Fig.~\ref{fig:detach} and only gradients from $R$ can back-propagate to previous layers. That is, the workload of learning two flows, to some extent, can be de-coupled in different time by gradually decreasing the TD rate to incorporate temporal information with representation features as training goes. To verify the effectiveness of this idea, several experiments are conducted in Section~\ref{sec:exp} and Section~\ref{sec:analysis}.
	
	\begin{figure}
		\begin{center}
			
			\includegraphics[width=0.8\linewidth, height=1.4in]{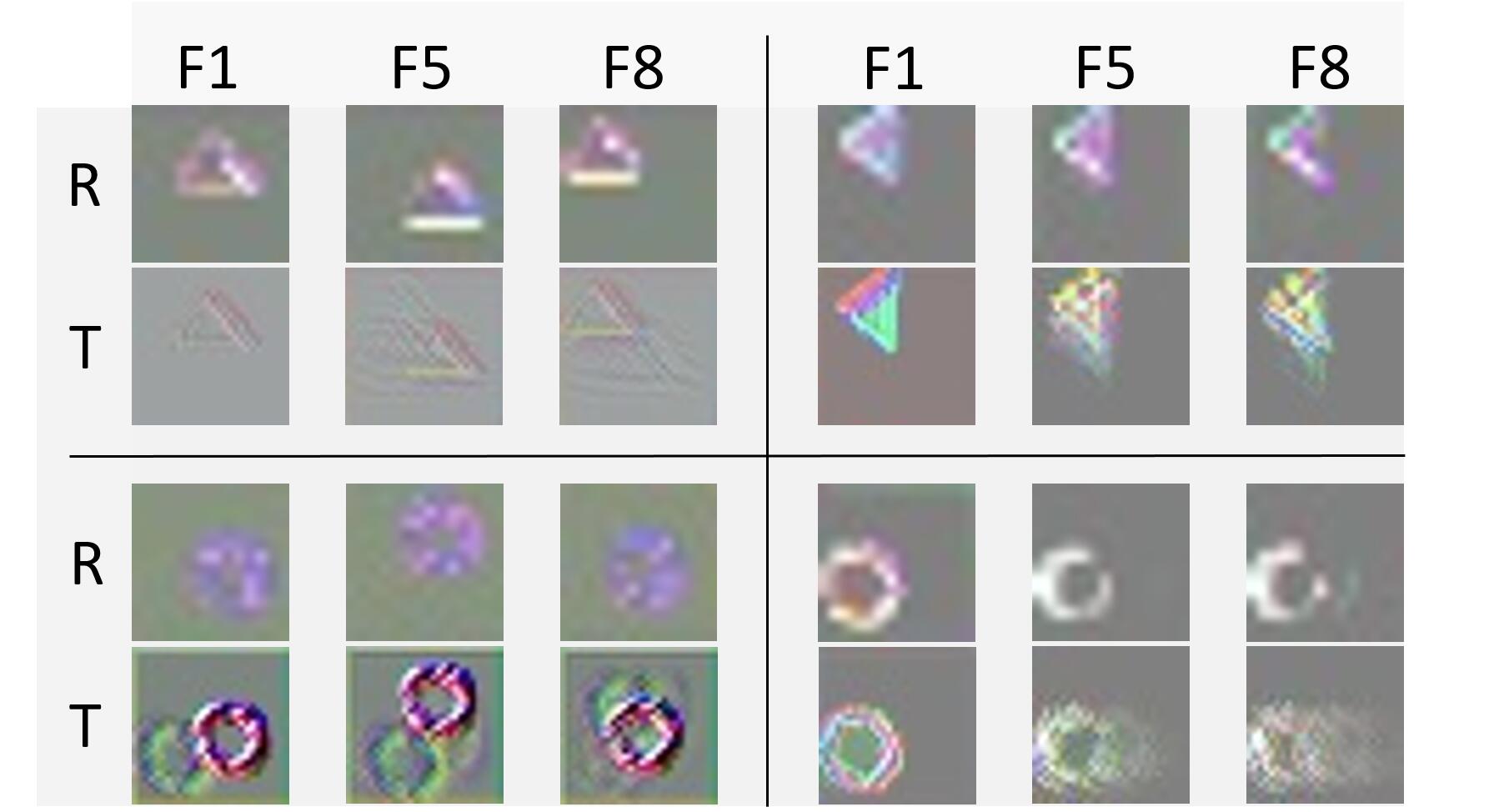}
		\end{center}
		\caption{Four examples of feature maps from the representation and temporal unit on the toy experiment. ``F1" denotes frame 1, ``R" denotes representation unit and ``T" denotes temporal unit.}
		\label{fig:featureMap}
		\vspace{-0.15in}
	\end{figure}

	\vspace{-0.15in}
	\paragraph{Comparison with Conventional RNN/LSTM}
	As mentioned before, stacked RNN/LSTM is a solution for deep recurrent architecture. Actually, our proposed approach is a general version of it. Specifically, when we set the output of $R$ unit as constant 1, our model degenerates into stacked RNN/LSTM model ($T$ unit can be LSTM cell). If we further set the depth of representation branch to 1, our model becomes a conventional shallow RNN/LSTM. From another perspective, our model can be considered as an extension of stacked RNN/LSTM with an extra context bridge, namely the $R$ unit.
	
	\vspace{-0.15in}
	\paragraph{Discussion with a Toy Experiment}
	To further provide an intuitive perception for the function of context bridge module, we design a toy experiment. The experiment is a video classification task that requires the model to learn which object is in the video from spatial information and how it moves from the temporal information, such as ``a triangle is moving left" or ``a circle is moving right". We adopt a 3-layer CBM model with 3 channels and visualize the feature maps of the final layer's representation unit and temporal unit in Fig.~\ref{fig:featureMap}. We can see the two computing units act as expected that the representation one mainly focuses on the spatial information while the temporal information is captured by the temporal unit.
	
	\begin{figure}
		\begin{center}
			\includegraphics[width=\linewidth, height=1.7in]{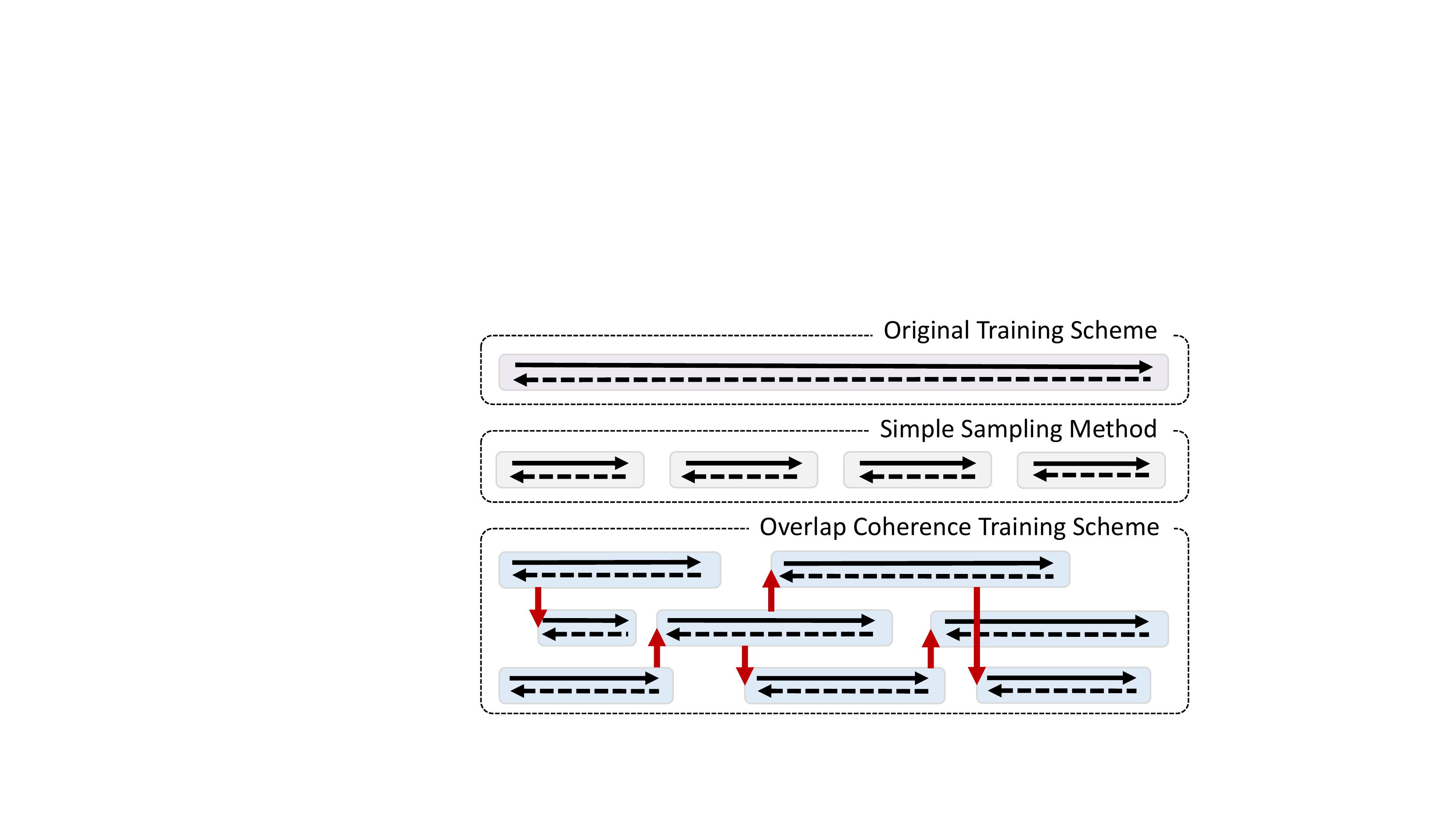}
		\end{center}
		\caption{Deep RNN training schemes. \textbf{First row:} original training scheme for RNN that takes the whole sequence as input and the information can flow forward and backward without interdicting. \textbf{Second row:} simple sampling method that samples the input long sequence into several short clips. \textbf{Third row:} our overlap coherence training scheme. Note that every item in the sequence can receive backward information (gradients) due to the existence of overlaps. The red line represents the initialization of each clip that is randomly chosen from the former clips with overlaps.}
		\label{fig:slice}
		\vspace{-0.15in}
	\end{figure}
	
	\subsection{Overlap Coherence Training Scheme}
	\paragraph{Challenge} In practice, utilizing deep RNN to model high-dimensional visual long sequences can be hard to achieve because with the depth increasing, the computing resources needed significantly expand. The deeper the model is, the more dramatically computational complexity grows with the increase of the sequence length, which can be regarded as a contradiction between the structural depth and sequence length (temporal depth). Recently, a widely-used method is to sample a few items from the long sequence (successive or scattered) and learn a truncated version of the original RNN on them~\cite{donahue2015long,hou2017tube,gu2017ava,carreira2017quo} to solve the contradiction. Under this scheme, training on short samples instead of the long sequence greatly reduces the training complexity, which is very practical for deep RNN. However, this can be seen as a compromise for the depth, which may lead to losing some key temporal information. Considering two short sampled clips that own overlaps, the outputs of the overlap sections must be different due to the broken temporal information, which will never happen if we train the whole sequence together, and this provides a clear evidence for the weakness of this sampling method.

	\vspace{-0.15in}
	\paragraph{Method} In this paper, we also consider shortening the long sequence to simplify the $l$-order Markov process into several $n$-order ($n < l$) ones, but we smooth the information propagation among short clips by introducing the \emph{Overlap Coherence Training Scheme}. In training phase, we randomly sample clips that have random lengths and starting points, which will naturally generate many overlaps (Third row in Fig.~\ref{fig:slice}). The overlaps serve as the communication bridge among the disconnected clips to make information flow forward and backward smoothly throughout the whole sequence. Therefore, we introduce a new loss called \emph{overlap coherence loss} to force the outputs of overlaps from different clips to be as closed (coherent) as possible. Then, the training objective function can be written as
	\vspace{-0.1in}
	\begin{equation}
	\sum_{i=1}^{N} \mathcal{L}_{r}(s_i) + \lambda \sum_{(v,u)\in \Omega } \mathcal{L}_d(v, u),
	\end{equation}
	where $s_i$ is the $i^{th}$ clip and $\Omega$ is the set of pairs which are the outputs of overlap sections from different clips. $\mathcal{L}_{r}$ and $\mathcal{L}_d$ denote the original loss for the specific task and our overlap coherence loss implemented by MSE loss respectively, where $\lambda$ is the hyper-parameter to adjust the weight of them.
	
	Additionally, our training scheme exhibits several highlights in practice. Firstly, our random sampling mode serves as a great data argumentation approach to enhance model's generalization ability. Secondly, the vanishing/exploding gradient problem can be solved to some extent since the scheme will shorten the sequence adequately to train easily. Thirdly, the initial state of each clip is taken from other earlier trained clips by picking up their hidden states at corresponding time stamp, which further bridges the information flow among clips to make it smoothly transfer throughout the whole sequence. Furthermore, initialized clips can be computed together in parallel, which can effectively reduce the training time, especially when the overlap rate is high.
	
\begin{figure}
		\begin{center}
			\includegraphics[width=\linewidth, height=0.55in]{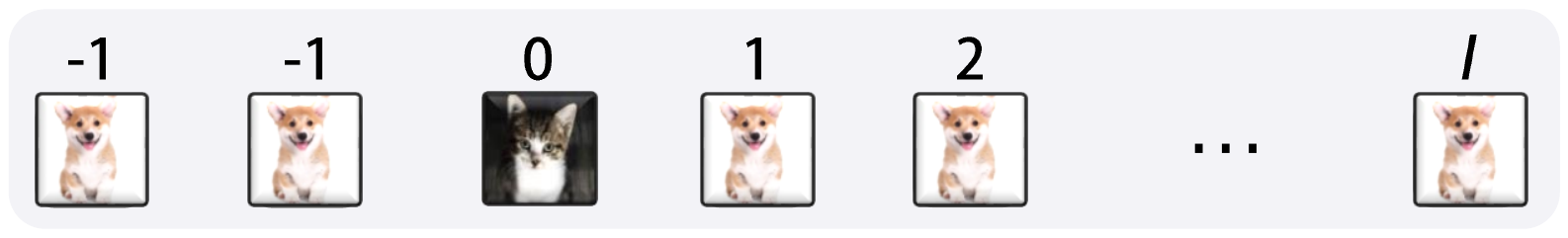}
		\end{center}
		\caption{``Cat \& Dog" experiment. The input sequence is images of cat and dog, and the label of each image represents the distance from the cat image (padding with -1).}
		\label{fig:b&w}
		\vspace{-0.15in}
	\end{figure}

	Moreover, to verify our training scheme can actually transfer useful information flow throughout the whole sequence, we commit a toy experiment shown in Fig.~\ref{fig:b&w}. The input sequence is a series of images, where there is only one cat and the others are all dogs. We train a model with overlap coherence training scheme to learn how far the current dog image is from the cat image appeared before. We find that the model can correctly predict even if the cat image appearing 50 frames ago, where we set the clip length smaller than 10. This is because temporal information of the image sequence is successfully captured among clips due to our overlap coherence training scheme.
	
	\vspace{-0.05in}
	\section{Experimental Results} \label{sec:exp}
	In this section, we evaluate our deep RNN framework and compare it with conventional shallow RNN (we choose the commonly used one: LSTM) on several sequence tasks to exhibit the superiority of our deep RNN framework over the shallow ones on high-dimensional inputs.
	
	\begin{figure}
		\begin{center}
			\includegraphics[width=\linewidth]{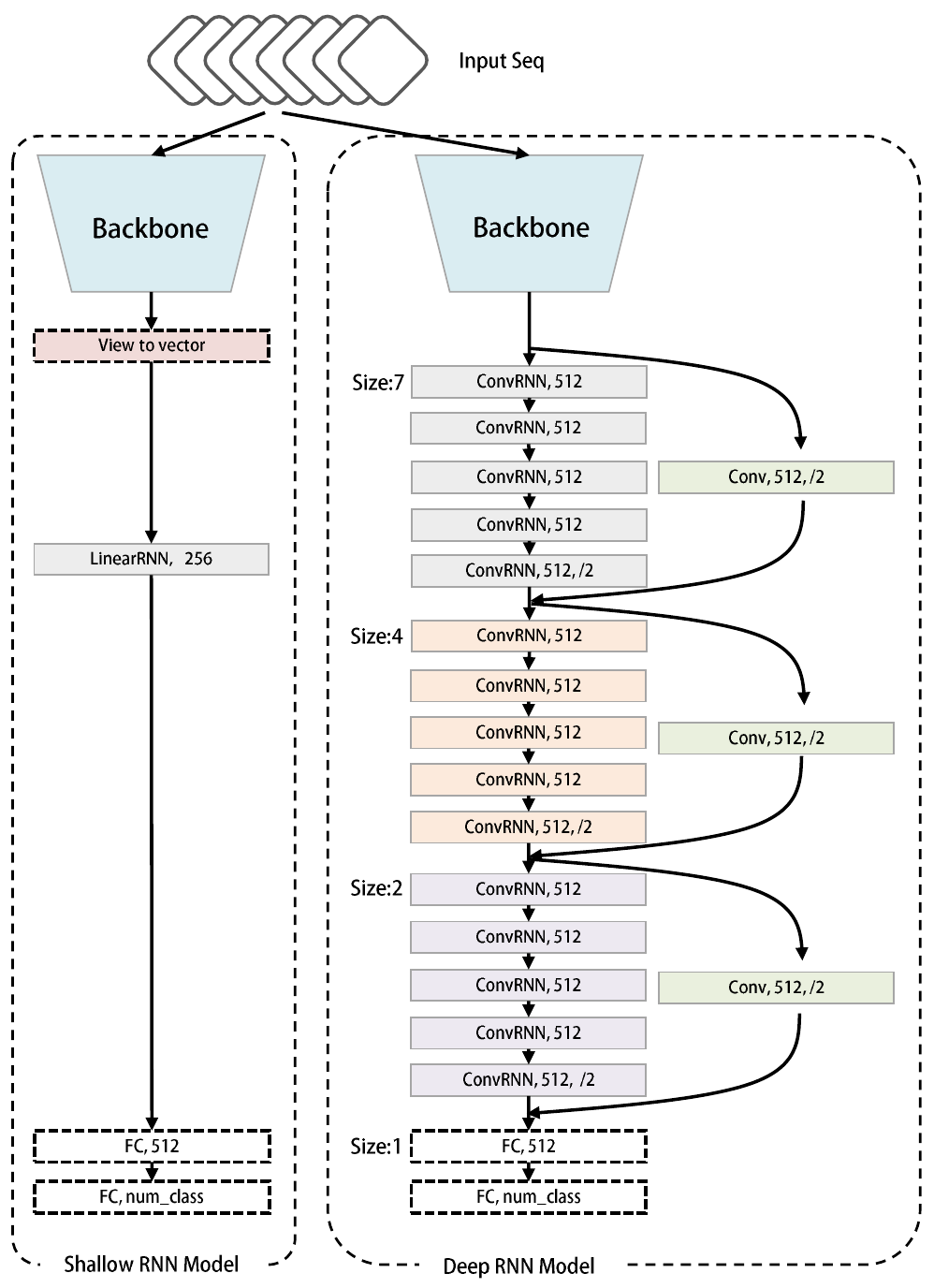}
		\end{center}
		\caption{Shallow and deep RNN architecture. The shallow version is implemented based on \cite{donahue2015long}. The deep one contains 15 RNN layers and we add shortcuts along the depth, following \cite{he2016deep}. Different from the shallow one, the RNN kernel is convolutional to maintain the spatial features instead of linear kernel.}
		\label{fig:models}
		\vspace{-0.15in}
	\end{figure}
	
	\subsection{Video Action Recognition and Anticipation}
	We first evaluate our method with action recognition and anticipation tasks~\cite{bian2017revisiting,long2018attention} on the UCF-101 dataset~\cite{soomro2012ucf101} and HMDB-51 dataset~\cite{kuehne2011hmdb} to compare our deep RNN with the common shallow one with CNN backbones. Then we remove the backbones, evaluate the standalone deep RNN model on Kinetics dataset~\cite{carreira2017quo} to compare it with several excellent approaches, not merely the shallow RNN.
	
	\vspace{-0.15in}
	\paragraph{Implementation}
	The frames in videos are resized with its shorter side into 368 and a $224\times224$ crop is randomly sampled from the frame or its horizontal flip. Color augmentation is used, where all the random augmentation parameters are shared among the frames of each video. We adopt BN~\cite{ioffe2015batch} after each convolutional layer, the same as ~\cite{ioffe2015batch}. The backbones (if needed) are pre-trained on ImageNet~\cite{russakovsky2015imagenet} and the RNN part is initialized by ``Xavier initialization" proposed in~\cite{glorot2010understanding}. We use Adam optimizer~\cite{kingma2014adam} with 64 mini-batch for shallow net and 16 for deep one. The learning rate starts from $10^{-4}$ and gradually decays. Besides, we adopt a weight decay of $10^{-5}$ and dropout of 0.2 and 0.5 for feature extractor and classifier respectively.

	\vspace{-0.11in}
	\paragraph{Adopting Conventional Backbone-Supported Structure}
	Conventional RNN model~\cite{donahue2015long} is stacking a 1-layer LSTM on the VGG~\cite{simonyan2014very} backbone. Now we extend it to deeper versions (shown in Fig.\ref{fig:models}) --- stacking a 15-layer ConvLSTM~\cite{xingjian2015convolutional} or a 15-layer CBM on the VGG backbone. For TD rate of CBM cell, we start from 1.0, decay to 0.8 after two epochs, and finally to 0.5 after another two epochs. We adopt our overlap coherence training scheme for both of the two deep versions to make them feasible, fix the weighting factor $\lambda=0.8$ for overlap coherence loss, and keep the overlap rate of sampling as 25\%.
	
	The results are shown in Tab.~\ref{tab:action}. For action recognition, the deep ConvLSTM model has a lower accuracy compared with the shallow model while for our deep CBM model, it obtains 12.2\% relative improvements on UCF-101 and 11.7\% on HMDB-51. For action anticipation, both of the two deep models gain improvements and our CBM version possesses the best performance --- achieving 88.6\% relative improvements on UCF-101 and 70.7\% on HMDB-51.
	
	Furthermore, we replace the VGG backbone with InceptionV1 to validate the universality of our deep RNN framework on UCF-101 of action recognition. Results are shown in Tab.~\ref{tab:inception}, where our deep CBM model still outperforms the shallow one, achieving 5.3\% relative improvements.
	\begin{table}[]
		\footnotesize
		\caption{Classification accuracy on UCF-101 and HMDB-51 (both on the first test split). For action recognition, the whole sequence is taken as input, while for anticipation, only the first two frames are used to do inference. Note that ``Recg" denotes action recognition task and ``Atcp" denotes action anticipation task.}
		\renewcommand{\arraystretch}{1.2}
		\begin{center}
			\begin{tabular}{c|c|c|c|c}
				\hline
				&\multicolumn{2}{c|}{UCF-101}&\multicolumn{2}{c}{HMDB-51}\\
				\cline{2-5}
				& Recg & Atcp & Recg & Atcp\\
				\hline
				1-layer LSTM & 71.1~\cite{donahue2015long} & 30.6~\cite{aliakbarian2017encouraging} & 36.0~\cite{carreira2017quo} & 18.8 \\
				15-layer ConvLSTM & 68.9 & 49.6 & 34.2 & 27.6\\
				\hline
				1-layer CBM & 65.3 & 28.4 & 34.3 & 16.9\\
				15-layer CBM & \textbf{79.8} & \textbf{57.7} & \textbf{40.2} & \textbf{32.1}\\
				\hline
			\end{tabular}
		\end{center}
		\label{tab:action}
		\vspace{-0.1in}
	\end{table}
	
	\begin{table}[]
		\footnotesize
		\caption{Action recognition accuracy on UCF-101 first test split. }
		\renewcommand{\arraystretch}{1.2}
		\begin{center}
			\begin{tabular}{c|c}
				\hline
				Model & Recognition Acc\\
				\hline
				1-layer LSTM with VGG ~\cite{donahue2015long} & 71.1 \\
				1-layer LSTM with InceptionV1~\cite{carreira2017quo} & 81.0\\
				15-layer ConvLSTM with InceptionV1& 77.6 \\
				15-layer CBM with InceptionV1& \textbf{85.3}\\
				\hline
			\end{tabular}
		\end{center}
		\label{tab:inception}
		\vspace{-0.2in}
	\end{table}
	
	\begin{table}[]
		\footnotesize
		\caption{Action recognition accuracy on Kinetics, and end-to-end fine-tuning on UCF-101 and HMDB-51. Note that our Deep CBM model applies 17 layers of CBM. ``BB" denotes backbone. 
		}
		\renewcommand{\arraystretch}{1.2}
		\begin{center}
			\begin{tabular}{c|c|c|c}
				\hline
				Architecture &  Kinetics & UCF-101 & HMDB-51\\
				\hline
				Shallow LSTM with BB~\cite{donahue2015long} & 53.9 & 86.8 & 49.7\\
				C3D~\cite{ji20133d} & 56.1 & 79.9 & 49.4\\
				Two-Stream~\cite{simonyan2014two} & 62.8 & 93.8 & 64.3\\
				3D-Fused~\cite{feichtenhofer2016convolutional} & 62.3 & 91.5 & 66.5\\
				Deep CBM without BB & 60.2 & 91.9  & 61.7 \\
				\hline
			\end{tabular}
		\end{center}
		\label{tab:kinetics}
		\vspace{-0.2in}
	\end{table}
	
	\vspace{-0.1in}
    %
    %
	
	\paragraph{Adopting Standalone RNN Structure}
	To reveal the excellent spatial representation ability of our deep RNN framework, we remove the backbone, adopt a standalone end-to-end deep RNN model to extract temporal and representation features simultaneously.
	
	Specifically, we utilize a deeper structure with 17-layer CBM, where the representation unit of each layer is set the same as the corresponding layer in ResNet-18~\cite{he2016deep} and the same shortcuts are employed. Other implementation details are consistent with the above backbone-supported version.
	
	The action recognition results on Kinetics-400 are shown in Tab.~\ref{tab:kinetics} and we also fine-tune the model on UCF-101 and HMDB-51. Compared with the conventional shallow LSTM with the backbone, our deep CBM achieves great improvements --- 5.9\% on UCF-101, 19.4\% on HMDB-51, 11.7\% on Kinetics, and the performance is competitive with some excellent non-recurrent models which are more powerful on this task.
	
	\subsection{Polygon-RNN on Cityscapes}
		\begin{table*}[]
		\caption{Video prediction results on KTH. ``T1" denotes the first frame to predict and ``Avg" denotes the average value.}
		\renewcommand{\arraystretch}{1.5}
		\begin{center}
			\resizebox{\textwidth}{1.4cm}{
				\fontsize{20pt}{20pt}\selectfont
				\begin{tabular}{c|c|c|c|c|c|c|c|c|c|c|c|c|c|c|c|c|c|c|c|c|c|c}
					\hline
					Method & Metric & T1 & T2 & T3 & T4 & T5 & T6 & T7 & T8 & T9 & T10 & T11 & T12 & T13 & T14 & T15 & T16 & T17 & T18 & T19 & T20 & Avg\\
					\hline
					\multirow{ 2}{*}{ConvLSTM~\cite{xingjian2015convolutional}} & PSNR & 33.8 & 30.6 & 28.8 & 27.6 & 26.9 & 26.3 & 26.0 & 25.7 & 25.3 & 25.0 & 24.8 & 24.5 & 24.2 & 23.7 & 23.2 & 22.7 & 22.1 & 21.8 & 21.7 & 21.6 & 25.3 \\
					& SSIM & 0.947 & 0.906 & 0.871 & 0.844 & 0.824 & 0.807 & 0.795 & 0.787 & 0.773 & 0.757 & 0.747 & 0.738 & 0.732 & 0.721 & 0.708 & 0.691 & 0.674 & 0.663 & 0.659 & 0.656 & 0.765 \\
					\hline 
					\multirow{ 2}{*}{MCnet~\cite{villegas2017decomposing}} & PSNR & 33.8 & 31.0 & 29.4 & 28.4 & 27.6 & 27.1 & 26.7 & 26.3 & 25.9 & 25.6 & 25.1 & 24.7 & 24.2 & 23.9 & 23.6 & 23.4 & 23.2 & 23.1 & 23.0 & 22.9 & 25.9 \\
					& SSIM & 0.947 & 0.917 & 0.889 & 0.869 & 0.854 & 0.840 & 0.828 & 0.817 & 0.808 & 0.797 & 0.788 & 0.779 & 0.770 & 0.760 & 0.752 & 0.744 & 0.736 & 0.730 & 0.726 & 0.723 & 0.804 \\
					\hline
					\multirow{ 2}{*}{Ours} & PSNR & 34.3 & 31.8 & 30.2 & 29.0 & 28.2 & 27.6 & 27.1 & 26.7 & 26.3 & 25.8 & 25.5 & 25.1 & 24.8 & 24.5 & 24.2 & 24.0 & 23.8 & 23.7 & 23.6 & 23.5 & \textbf{26.5}\\
					& SSIM & 0.951 & 0.923 & 0.905 & 0.885 & 0.871 & 0.856 & 0.843 & 0.833 & 0.824 & 0.814 & 0.805 & 0.796 & 0.790 & 0.783 & 0.779 & 0.775 & 0.770 & 0.765 & 0.761 & 0.757 & \textbf{0.824}\\
					
					\hline
				\end{tabular}
			}
		\end{center}
		\label{tab:pred}
		\vspace{-0.2in}
	\end{table*}
	
	\begin{table}[]
		\footnotesize
		\renewcommand{\arraystretch}{1.2}
		\caption{Structures of Polygon-RNN models with different depths.}
		\begin{center}
			\begin{tabular}{c|c|c|c|c|c|c}
				\hline
				\multicolumn{2}{c|}{\# filters} & 256 & 128 & 64 & 32 & 8\\
				\hline
				\multirow{4}{*}{\# layers} & 2-layer model& - & - & 1 & - & 1\\
				& 5-layer model & - & 2 & 1 & 1 & 1\\
				& 10-layer model & - & 5 & 3 & 1 & 1\\
				& 15-layer model & 2 & 4 & 6 & 2 & 1\\
				\hline
			\end{tabular}
		\end{center}
		\label{tab:polygonStructure}
		\vspace{-0.2in}
	\end{table}

	For auxiliary annotation task, similar with instance segmentation task~\cite{xu2018srda}, we build the model following Polygon-RNN~\cite{castrejon2017annotating}, and evaluate it on Cityscapes instance segmentation dataset~\cite{cordts2016cityscapes} which contains eight object categories and we use the same train/test split as \cite{castrejon2017annotating}.
	
	To build our model, we only replace the RNN part in the original Polygon-RNN model with our deep RNN framework which is a plain stacking of our CBM cell or ConvLSTM~\cite{xingjian2015convolutional} cell. Unlike the deep architecture shown in Fig.~\ref{fig:models}, we do not use shortcuts in this experiment. Inside the CBM cell, we still choose the element-wise production as merge function and set the size of all convolutional kernels as $3\times3$. For TD rate, we start from 1.0, decay to 0.8 after the first epoch, and finally to 0.5 after another one epoch. We evaluate our deep RNN framework with different layers, and Tab.~\ref{tab:polygonStructure} summarizes the specific architectures.
	
	\vspace{-0.15in}
	\paragraph{Implementation}The size of input images is $ 224\times224$. We adopt BN~\cite{ioffe2015batch} but with no dropout~\cite{hinton2012improving}. We initialize the convolutional layers with ``Xavier initialization"~\cite{glorot2010understanding}. Models are trained with a mini-batch size of 16 using Adam optimizer ~\cite{kingma2014adam}, and the learning rate starts from $10^{-4}$ and gradually decays when meeting the loss plateaus. We train deep models (10 and 15-layer ones) with the overlap coherence training scheme, where we set and keep $\lambda=0.8$.

	\vspace{-0.15in}
	\paragraph{Results}We compare the 2, 5, 10, 15-layer RNN networks with ConvLSTM or CBM cell. The results are shown in Tab.~\ref{tab:polygonResult}. Compared with the original Polygon-RNN with the shallow RNN, our deep CBM model achieves 14.7\% relative improvements which is even competitive with Polygon-RNN++ proposed in~\cite{acuna2018efficient} which adopts many complex tricks, while the deep ConvLSTM model suffers from higher training loss, leading to a bad performance. 
	
	\subsection{Video Future Prediction}
	
	For video future prediction, we evaluate our deep RNN framework using the state-of-the-art method: MCnet proposed in~\cite{villegas2017decomposing}, which predicts 20 future frames based on the observed 10 previous frames. We only replace the 1-layer ConvLSTM part of the motion encoder into our 15-layer deep CBM model, where the TD rate is finally set to 0.5 with similar process as the above. The detailed implementation settings are consistent with the original method in~\cite{villegas2017decomposing}.
	
	We evaluate on the KTH dataset~\cite{schuldt2004recognizing} which contains 600 videos for 6 human actions, and we utilize PSNR~\cite{mathieu2015psrn} and SSIM~\cite{wang2004ssim} as metrics. The results are shown in Tab.~\ref{tab:pred} and we can see that compared with the original method using shallow RNN, our deep model achieves 1.6\% improvements on SSIM and 1.8\% on PSNR for 10-frame prediction, and 2.6\% on SSIM and 2.1\% on PSNR for 20-frame prediction. 
	
	In this experiment, we do not adopt the overlap coherence training scheme since the sequence is not too long. 
	
	
	\begin{table}[]
		
		\caption{Performance (IoU in \%) on Cityscapes validation set (used as test set in~\cite{castrejon2017annotating}). Note that ``Polyg-LSTM" denotes the original Polygon-RNN structure with ConvLSTM cell and ``Polyg-CBM" denotes the Polygon-RNN structure with CBM cell.}
		\renewcommand{\arraystretch}{1.2}
		\begin{center}
			\footnotesize
			\begin{tabular}{c|c|c|c}
				\hline
				\multicolumn{3}{c|}{Model} & IoU\\
				\hline
				\multicolumn{3}{c|}{Original Polygon-RNN~\cite{castrejon2017annotating}} & 61.4\\
				\multicolumn{3}{c|}{Residual Polygon-RNN~\cite{acuna2018efficient}} & 62.2\\
				\multicolumn{3}{c|}{Residual Polygon-RNN + attention + RL~\cite{acuna2018efficient}} & 67.2\\
				\multicolumn{3}{c|}{Residual Polygon-RNN + attention + RL + EN~\cite{acuna2018efficient}} & 70.2\\
				\multicolumn{3}{c|}{Polygon-RNN++~\cite{acuna2018efficient}} & \textbf{71.4}\\
				\hline
				& \# layers & \# params of RNN & \\
				\hline
				Polyg-LSTM & 2 & 0.47M & 61.4\\
				Polyg-LSTM & 5 & 2.94M & 63.0\\
				Polyg-LSTM & 10 & 7.07M & 59.3\\
				Polyg-LSTM & 15 & 15.71M & 46.7\\
				\hline
				Polyg-CBM & 2 & 0.20M & 59.9\\
				Polyg-CBM & 5 & 1.13M & 63.1 \\
				Polyg-CBM & 10 & 2.68M & 67.1\\
				Polyg-CBM & 15 & 5.85M & \textbf{70.4}\\
				\hline
			\end{tabular}
		\end{center}
		\label{tab:polygonResult}
		\vspace{-0.2in}
	\end{table}
		
	\vspace{-0.05in}
	\section{Analysis}  \label{sec:analysis}
	The above visual applications demonstrate the superiority of our deep RNN framework and in this section we will further verify the effectiveness of our detailed designs --- the model depth, CBM for deep structure, the overlap coherence training scheme, merge function and TD rate of CBM.
	
	\begin{figure*}
		\centering
		\subfigure[Polyg-LSTM]{
			\label{fig:polyg-fig:a} 
			\includegraphics[width=0.34\linewidth, height=1.1in]{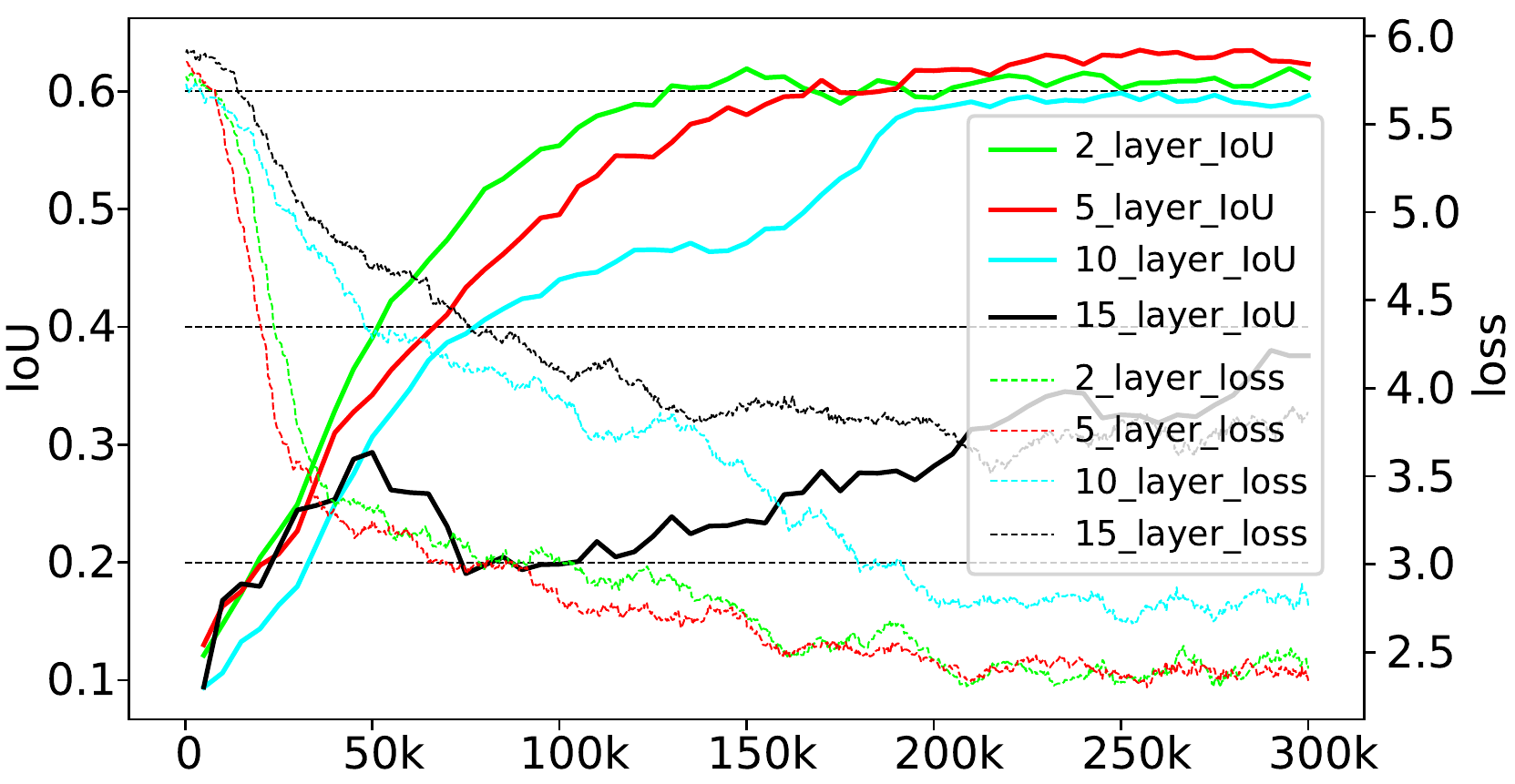}}
		\subfigure[Polyg-CBM]{
			\label{fig:polyg-fig:b} 
			\includegraphics[width=0.34\linewidth, height=1.1in]{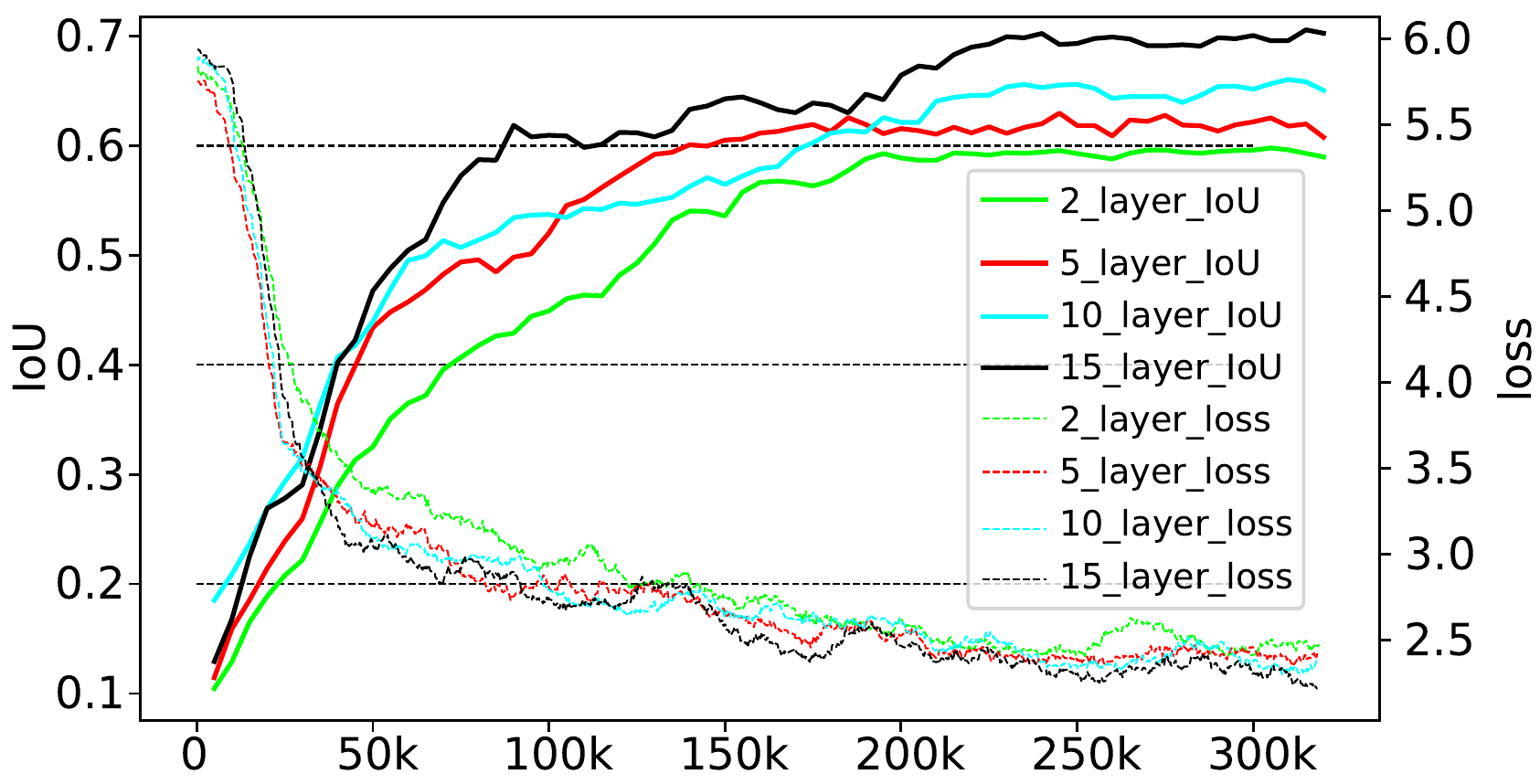}}
		\subfigure[TD-rate comparison]{
			\label{fig:polyg-fig:c} 
			\includegraphics[width=0.29\linewidth, height=1.1in]{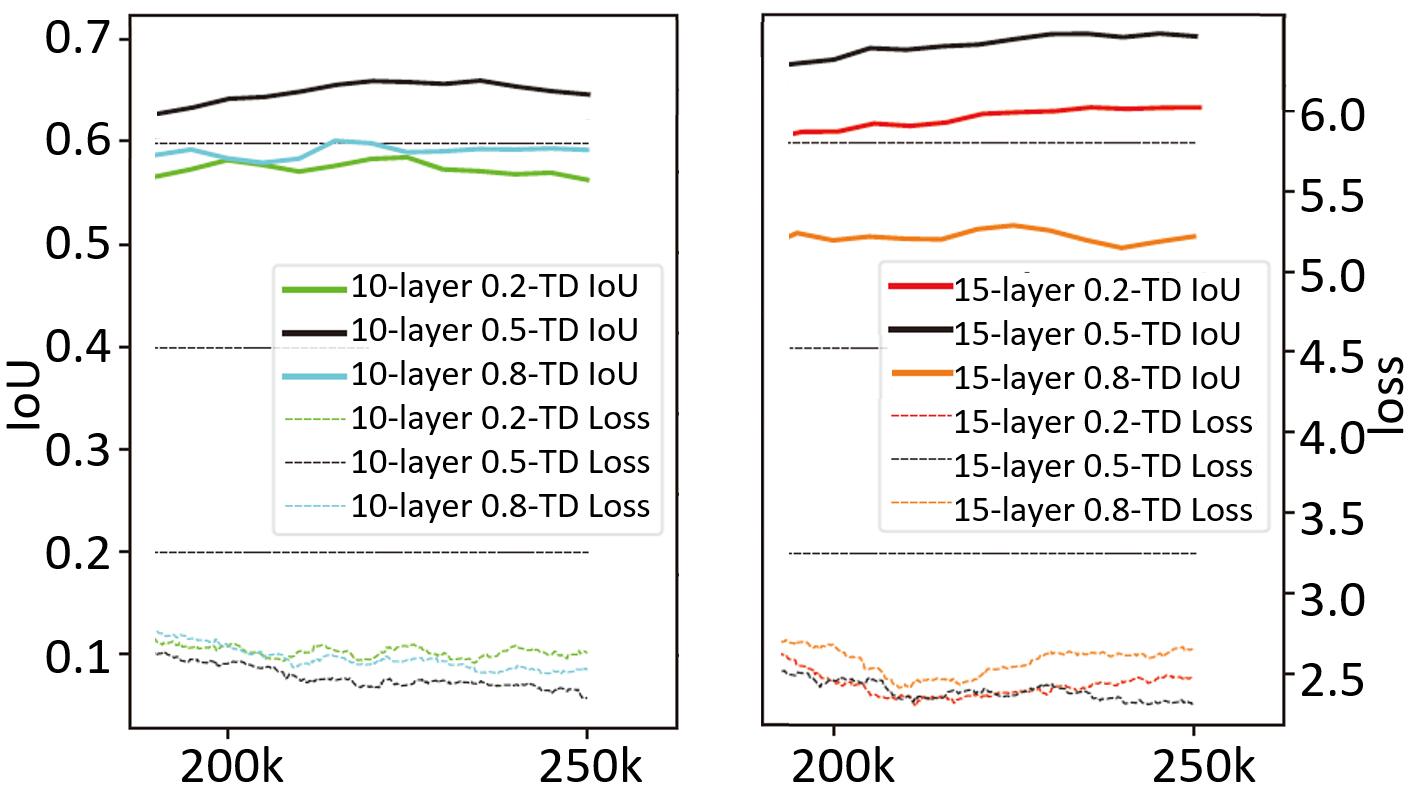}}
		
		\caption{Training on Cityscapes. Dashed lines denote training loss, and the bold lines denote testing IoU. \textbf{Left:} Polyg-LSTM networks. Deep models are difficult to train and suffer from high training loss. The convergence of 15-layer is not shown. \textbf{Middle:} Polyg-CBM networks adopting 0.5 TD-rate. Deep models are easy to train. \textbf{Right:} Comparison between different TD rates on 10 and 15-layer models.}
		\label{fig:polyg-fig} 
		\vspace{-0.15in}
	\end{figure*}

	\vspace{-0.15in}
	\paragraph{Analysis on Depth}
	Results of all above experiments have already demonstrated that our deep RNN model remarkably outperforms the shallow RNN one due to the stronger representation capability with the depth growing. We analyze the experiments on Polygon-RNN to further explore the specific relationships between the depth and the model performance, which is illustrated in Fig.~\ref{fig:polyg-fig:a} and Fig.~\ref{fig:polyg-fig:b}. 
	
	From Fig.~\ref{fig:polyg-fig:b}, we can observe that utilizing CBM, the deeper the model is built, the lower training loss and higher IoU value we will receive. Moreover, it is worth noting that the deep models converge as fast as the shallow ones.
	
	

	\vspace{-0.15in}
	\paragraph{Analysis on CBM}
	As results shown in Tab.~\ref{tab:action} and Tab.~\ref{tab:inception}, our deep CBM model achieves the best performance on action recognition task with two different backbones and action anticipation task, while deep ConvLSTM model suffers from lower accuracy on action recognition even compared with the shallow one. 
	
	As we discussed above, building deep RNN models needs to co-adapt to both temporal and representation information, making it difficult to optimize over a long sequence. Therefore, for action recognition that takes the whole videos as inputs, commonly-used deep RNN models cannot benefit from the increased depth, while for action anticipation that predicts only based on the first two frames, deeper structure brings better results. To resolve this problem caused by the contradiction of the two information flows when stacking deep, our CBM cell is right introduced to de-couple these two flows to make training more efficient, and receives best results on both tasks.

	Besides, results of Polygon-RNN task in Tab.~\ref{tab:polygonResult} also prove that our CBM cell is more suitable for stacking deep, and comparisons between Fig.~\ref{fig:polyg-fig:a} and Fig.~\ref{fig:polyg-fig:b} further reveal that using ConvLSTM to stack deep leads to higher training loss and lower IoU value.
	
	\vspace{-0.15in}
	\paragraph{Analysis on Overlap Coherence Training Scheme}
	All the deep models above adopt our overlap coherence training scheme. From the results, we can see that it works well --- deep models are trainable on commonly-used GPUs and all the models learn effective temporal features. Under this scheme, though it may not transfer temporal information as smoothly as the original training scheme, the overlaps and the coherence loss guarantee the consistency of temporal information among the clips to a certain degree, and finally we do benefit from the increasing structural depth by making some compromise on the sequence length.
	
	\begin{table}[]
		\footnotesize
		\caption{Classification accuracy on UCF-101 with element-wise production and addition settings. For $R$, both of the two settings adopt $ReLU(Conv(\cdot))$. For $T$, production setting adopts $Sigmoid(Conv(\cdot))$ while addition adopts $ReLU(Conv(\cdot))$.}
		\renewcommand{\arraystretch}{1.2}
		\begin{center}
			\begin{tabular}{c|c|c}
				\hline
				& Recognition & Anticipation\\
				\hline
				Production & 79.8 & 57.7\\
				Addition & 77.4 & 56.7\\
				\hline
			\end{tabular}
		\end{center}
		\label{tab:mergeFunc}
		\vspace{-0.2in}
	\end{table}
	
	\vspace{-0.15in}
	\paragraph{Analysis on Merge Function $\zeta$}
	All above experiments are committed with element-wise production merge function. Here, we also evaluate another setting: $ReLU(Conv(\cdot))$ for $R$ and $T$, and element-wise addition for the merge function, which treats the two flows equally without discrimination when merging the information. For action recognition and anticipation tasks on UCF-101, the comparison of these two settings is shown in Tab.~\ref{tab:mergeFunc}. We find that the production setting is marginally better than the addition one, possibly because the production setting extracts better spatial representation features that are more useful for video classification problems.
	
	\vspace{-0.15in}
	\paragraph{Analysis on TD Rate}
	To show the influence of TD rate, we set the final TD rate to 0.0, 0.2, 0.5, 0.8 and 1.0 (gradually decay as the above experiments) and results of action recognition task on UCF-101 are shown in Tab.~\ref{tab:detchout}. We can see that 0.5 TD-rate achieves the best result. When the TD rate is set to 1.0, the temporal information can only flow backward in its own layer, forbidding the temporal communication among different layers, thus leading to a relatively non-ideal performance. For Polygon-RNN task, the results shown in Fig.~\ref{fig:polyg-fig:c} reveal consistent conclusions.
	\begin{table}[]

		\footnotesize
		\caption{Action recognition accuracy on UCF-101 with different TD rates. We use VGG19 as backbones, 15-layer CBM as the RNN part, and element-wise production as merge function. }
		\renewcommand{\arraystretch}{1.2}
		\begin{center}
			\begin{tabular}{c|c||c|c||c|c}
				\hline
				TD rate &  Acc & TD rate & Acc & TD rate & Acc\\
				\hline
				0.0 & 75.2 & 0.5 & \textbf{79.8} & 1.0 & 75.3\\
				0.2 & 76.5 & 0.8 & 77.1 &&\\
				\hline
			\end{tabular}
		\end{center}
		\label{tab:detchout}
		\vspace{-0.2in}
	\end{table}
	
	\vspace{-0.1in}
	\section{Conclusion}
	In this paper, we proposed a deep RNN framework for visual sequential applications. 
	The first part of our deep RNN framework is the CBM structure designed to balance the temporal flow and representation flow. Based on the characteristics of these two flows, we proposed the Temporal Dropout to simplify the training process and enhance the generalization ability.
	The second part is the Overlap Coherence Training Scheme aiming at resolving the large resource consuming of deep RNN models, which can significantly reduce the length of sequences loaded into the model and guarantee the consistency of temporal information through overlaps simultaneously.  
	
	We conducted extensive experiments to evaluate our deep RNN framework. Compared with the conventional shallow RNN, our deep RNN framework achieves remarkable improvements on action recognition, action anticipation, auxiliary annotation and video future prediction tasks. Comprehensive analysis is presented to further validate the effectiveness and robustness of our specific designs.
	
	\vspace{-0.1in}
	\section{Acknowledgements}
	This work is supported in part by the National Key R\&D Program of China, No. 2017YFA0700800, National Natural Science Foundation of China under Grants 61772332.

	{\small
		\bibliographystyle{ieee}
		\bibliography{egbib}
	}

%

\end{document}